\documentclass[10pt,twocolumn,letterpaper]{article}
\usepackage{times}
\usepackage{helvet}
\usepackage{courier}
\usepackage[T1]{fontenc}
\usepackage[hyphens]{url}
\usepackage{graphicx}
\usepackage{natbib}
\usepackage{caption}
\usepackage{enumitem}
\usepackage{float}
\setlist[itemize]{noitemsep,topsep=0pt,parsep=0pt,partopsep=0pt}
\usepackage{algorithm}
\usepackage{algorithmic}
\usepackage{newfloat}
\usepackage{listings}
\renewcommand{\arraystretch}{1.18}
\DeclareCaptionStyle{ruled}{labelfont=normalfont,labelsep=colon,strut=off}
\floatstyle{ruled}
\newfloat{listing}{tb}{lst}{}
\floatname{listing}{Listing}

\usepackage[hidelinks]{hyperref}
\begin{document}

\title{
\vspace{-1.5em}
\includegraphics[width=0.18\textwidth]{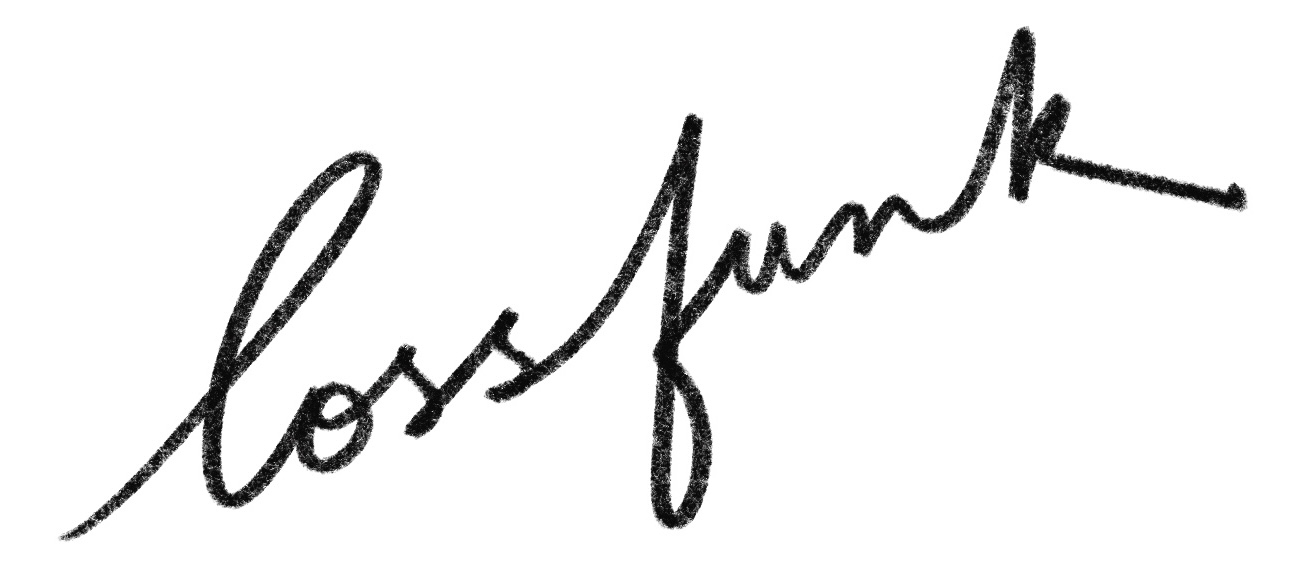}\\[0.8em]
Language Models Entangle Language and Culture
}

\author{
  Shourya Jain$^{1}$, Paras Chopra$^{1}$ \\
  $^{1}$Lossfunk \\
  \texttt{shourya.jain@lossfunk.com, paras@lossfunk.com}
}

\date{}

\maketitle

\begin{abstract}
Users should not be systemically disadvantaged by the language they use for interacting with LLMs; i.e. users across languages should get responses of similar quality irrespective of language used. In this work, we create a set of real-world open-ended questions based on our analysis of the WildChat dataset and use it to evaluate whether responses vary by language, specifically, whether answer quality depends on the language used to query the model. We also investigate how language and culture are entangled in LLMs such that choice of language changes the cultural information and context used in the response by using LLM-as-a-Judge to identify the cultural context present in responses. To further investigate this, we evaluate LLMs on a translated subset of the CulturalBench benchmark across multiple languages. Our evaluations reveal that LLMs consistently provide lower quality answers to open-ended questions in low resource languages. We find that language significantly impacts the cultural context used by the model. This difference in context impacts the quality of the downstream answer.
\end{abstract}

\section{Introduction}
Large Language Models (LLMs) such as ChatGPT \cite{openai2025chatgpt}, are used by hundreds of millions of people for their day-to-day queries. People ask LLMs for advice on a wide variety of topics such as healthcare, finance, education, etc., and the responses impact their decision making. Users use a variety of languages to interact with LLMs. LLMs are expected to provide advice/responses of the same quality across languages, i.e., switching the language of the query should not affect the quality of advice received otherwise, it can significantly impact the decision making and put users interacting with LLMs in a particular language at a disadvantage. Current work on evaluating LLMs in multilingual context mostly focuses on general knowledge, instruction-following, mathematical or programming capabilities, etc. Several multilingual LLM benchmarks and evaluations focus on niche domains. These benchmarks fail to consider the changes in style and cultural context in LLM responses introduced by the changes in the language used for interacting with these LLMs. Various studies evaluating bias in LLMs use cultural cues like name, nationality and ethnicity in queries which does not reflect how users frame their queries. There is a gap for evaluating the multilingual capabilities of LLMs on generic queries that people ask these LLMs on a day-to-day basis.

We fill this gap by creating generic advice seeking questions based on our analysis of the WildChat dataset \cite{zhao2024wildchat1mchatgptinteraction} and evaluating a set of multilingual LLMs on these questions across languages. We evaluate differences in answer quality by using an LLM as a judge to score responses. We cover a wide variety of model families for our evaluation: Qwen3 \cite{yang2025qwen3technicalreport}, Magistral \cite{mistralai2025magistral}, Sarvam-m \cite{sarvam2025blog} and Cohere-Aya \cite{dang2024ayaexpansecombiningresearch}, which were specifically trained for multilingual use. We evaluate LLM performance across English, Hindi, Chinese, Swahili, Hebrew, and Brazilian Portuguese. To ensure the quality of scores, we use Cohere Command-A \cite{cohere2025CommandA} as the judge model due to its advanced multilingual capabilities. We conduct an experiment to verify the absence of language bias in the judge model, the details for which can be found in subsection ~\ref{sec:judge-verification}. 

To further investigate the related nature of language and culture, we use LLM as a Judge to evaluate if asking the same query in different languages lead to culturally different answers. Our findings reveal that even for generic advice seeking questions, prompting in different languages leads to qualitatively and culturally different answers. To validate the entangled nature of language and culture, we use CulturalBench \cite{chiu2025culturalbenchrobustdiversechallenging}: a benchmark which tests cultural factual knowledge. It has 1696 factual knowledge questions spanning 45 geographic regions. We take a subset of this benchmark with more than 750 questions from 29 regions, and  create a translated version to evaluate LLM performance on the benchmark across languages. Our findings indicate that the performance of LLMs on factual questions related to any geographical location varies significantly across languages.

We attribute this difference in answers across languages to language and culture being related for LLMs, causing LLMs to use different cultural information when the same query is asked in different languages. 

Our contributions can be summarized as:
\begin{itemize}
 \item We create a set of generic advice-seeking queries based on our analysis of the WildChat dataset.
 \item To the best of our knowledge, this work presents the first qualitative evaluation of LLM responses across languages for generic, advice-seeking queries.
 \item We create and (will) release a translated version of the CulturalBench dataset.
 \item We demonstrate that language and culture are entangled in LLMs such that the choice of language used in the query impacts the cultural context used in the response.
\end{itemize}

\begin{figure*}[t]
\centering
\includegraphics[width=\textwidth]{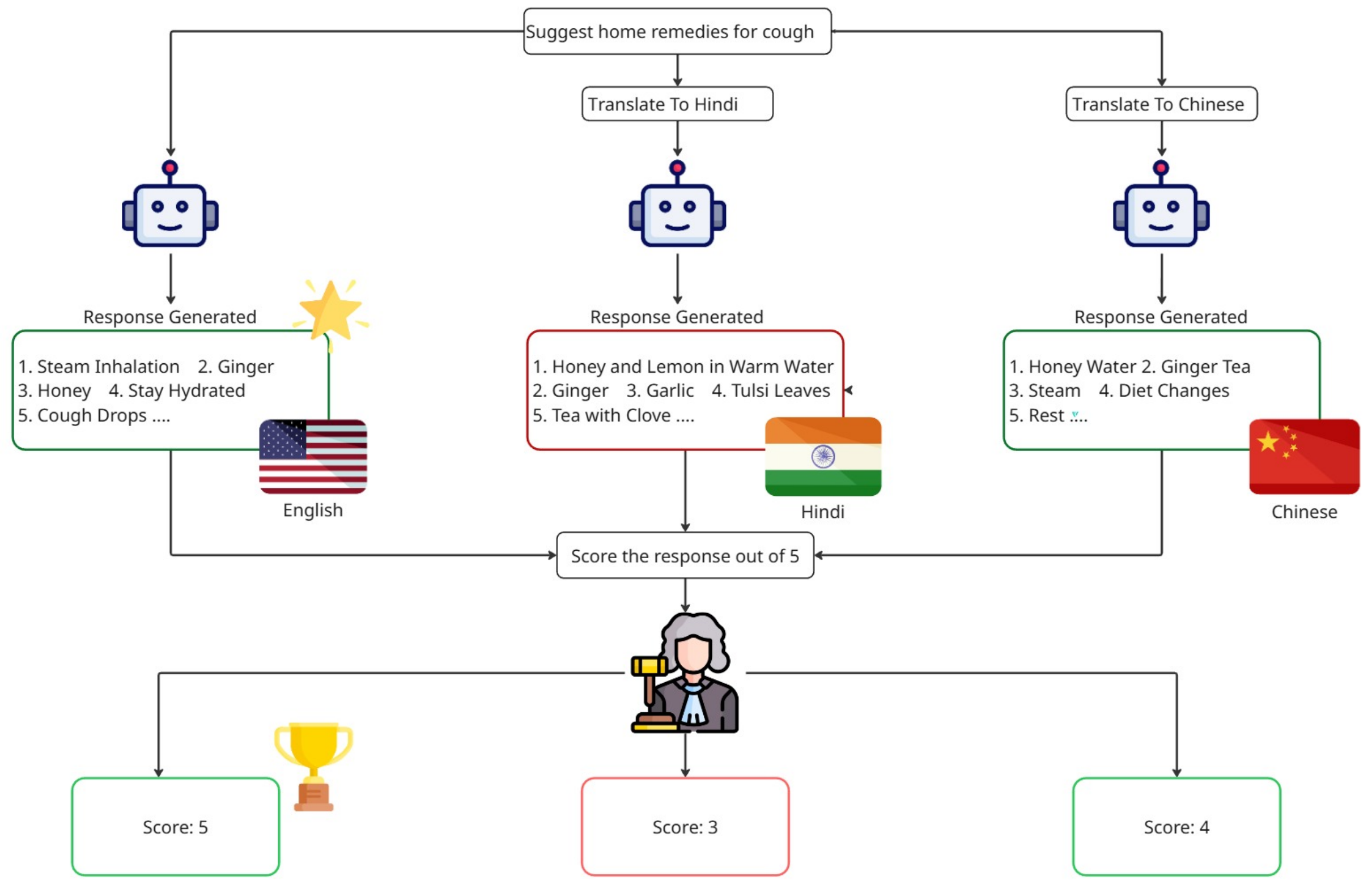}
\caption{\textbf{We show an example of our evaluation methodology. (1) Each query is translated to multiple languages. (2) We provide the translated and original English query to a LLM and a response is generated for each language. (3) Responses are scored out of 5 using a LLM as a Judge. (We show the responses translated to English for the visualization. Responses are evaluated in the original language itself.) }}
\label{fig:mechanism}
\end{figure*}

\section{Related Work}
\paragraph{Multilingual Evaluations:} Most multilingual benchmarks like MMMLU \cite{hendrycks2021measuringmassivemultitasklanguage}, MMLU-ProX \cite{xuan2025mmluproxmultilingualbenchmarkadvanced}, BenchMAX \cite{huang2025benchmaxcomprehensivemultilingualevaluation} evaluate LLM multilingual capabilities on instruction following, general knowledge, general reasoning etc. Other benchmarks like MGSM \cite{shi2022languagemodelsmultilingualchainofthought}, MSVAMP \cite{chen2023breaking}, Polymath \cite{wang2025polymathevaluatingmathematicalreasoning} focus on mathematical reasoning. These benchmarks mostly evaluate LLMs on multiple-choice questions (MCQs) or short form answers. Such evaluations fail to consider the stylistic and cultural variations in answers across languages and only focus solely on accuracy. Other works, such as INDIC QA \cite{singh2025indicqa}, xSquaD \cite{Artetxe_2020} evaluate multilingual question-answering performance given a context passage and a question. 

\paragraph{}Existing work on multilingual bias in LLMs uses prompts with cultural cues in the form of name, race, gender, nationality, country of residence etc. Such work includes \cite{rodríguez2025colombianwaitressesyjueces}, \cite{devinney2024dont}. Such works study bias along one or more axes such as gender, religion, race etc. While these evaluations are useful, most users do not necessarily use similar cues when interacting with LLMs. We differ from such studies as we use culture neutral prompting and do not study bias on any predefined axis. Other works study multilingual bias in narrower domains or tasks such as \cite{Bak2025crosslinguistic}, which evaluates bias in writing e-mails across languages and \cite{schlicht2025disparitiesmultilingualllmbasedhealthcare} examines bias in healthcare related queries. Overall, these evaluations are niche and do not consider the broader variety of queries that users commonly ask LLMs. Our work considers a broader set of queries based on our analysis of WildChat dataset.
\paragraph{Cultural Bias:} Current studies of multilingual cultural bias in LLMs use human cultural value surveys such as WVS \cite{WVS2022} or EVS \cite{EVS2022}. Works such as \cite{rystrøm2025multilingualmulticulturalevaluating, sukiennik2025evaluationculturalvaluealignment, Tao_2024} compare model responses across languages to human cultural values from survey data. Other studies such as \cite{aksoy2024moralityspeakunravelingcultural} analyses LLM responses across languages on MFQ-2 morality questionnaire \cite{Atari2023MoralityBeyondWEIRD}. These works mostly evaluate LLM choices on MCQs, analyze responses on Likert scales, or consider short responses. Such evaluations overlook aspects of culture such as history, cuisine, and etiquette. Our work differs by using open-ended queries and cultural-knowledge benchmarks to evaluate the relationship between language and culture. Our work is novel in its use of culture-neutral prompts and in evaluating performance on generic, open-ended queries without predefined bias axes. IndQA \cite{OpenAI2025IndQA} is a similar work focusing on Indian languages for evaluating multilingual LLM responses for queries requiring cultural knowledge. Our work is broader as it covers languages from various regions and shows how cultural differences can be present in multilingual responses even when queries do not require cultural context. While several work studies language and cultural biases, there is no work establishing a relation between language and culture to the best of our knowledge.

\begin{figure*}[t]
\centering
\includegraphics[width=\textwidth]{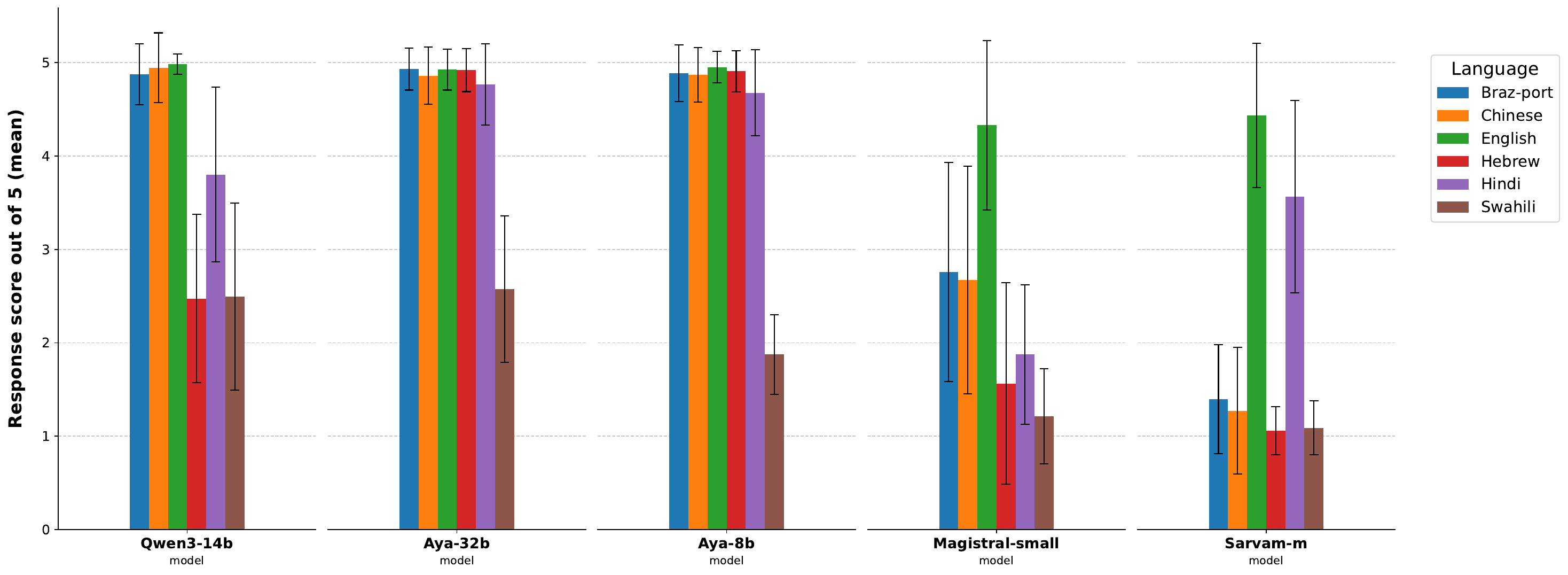}
\caption{\textbf{Comparison of answer quality across languages by model evaluated using LLM as a Judge. The results show all models provide worse responses in at least one language and all models show the best performance in English.}}
\label{fig:model_scores}
\end{figure*}

\section{Methodology}
\subsection{Question Generation}
\label{question_generation}
We create a set of 20 advice seeking questions covering a wide variety of topics like Healthcare, Business, Education. The complete list of questions and categories can be found in Appendix ~\ref{sec:questions}. These questions were created based on our analysis of the WildChat Dataset. As part of our analysis, we begin with initial filtering and cleaning. From the WildChat dataset, we retain only queries in English. We then remove queries related to programming bugs or error fixes. We found that such queries dominated the user queries, but we exclude them from our evaluations because they are asked primarily by a niche subset of users whose high frequency skews the dataset. We only keep queries with lengths ranging from 40 to 400 characters and exclude duplicate or highly similar queries with a threshold of 60 using the \textit{fuzzywuzzy} library \cite{fuzzywuzzy}. We converted the queries to embeddings using Qwen3-0.6b embedding model \cite{zhang2025qwen3embeddingadvancingtext}. We clustered the queries using the HDBSCAN algorithm \cite{campello2013hdbscan} followed by manual analysis of queries for creating the queries used for evaluation. The questions were structured in a culture-independent manner such that no culture related information is present in any of the queries. We translated the queries to Chinese, Hindi, Brazilian Portuguese, Swahili and Hebrew using Gemini-2.5-Flash model with temperature set to 0.

\subsection{Models Evaluated}
Our evaluation covered the following models: Qwen3-14B, Cohere-Aya-32B, Cohere-Aya-8B, Magistral and Sarvam-m. We selected these models to represent a variety of providers. Qwen3-14B is from Qwen (a Chinese provider); Cohere-Aya models are from Cohere (a Canadian provider) and were trained specifically for multilingual use cases; Magistral was developed by Mistral (a French provider); and Sarvam-m is a finetune over Mistral-Small tailored for the Indian use case. The models were evaluated via API using OpenRouter, except for Cohere models and Sarvam-m, for which we used the providers' respective
API platforms. Figure ~\ref{fig:mechanism} shows our evaluation mechanism.

\subsection{Evaluation Methodology}
To ensure the quality of evaluation using LLM-as-a-Judge, we performed several ablations to choose the best configuration. We took a subset of 10 queries from the 20 queries we created in subsection~\ref{question_generation} and a subset of languages: English, Hindi, Chinese and Hebrew. For each query and language pair, we prompted Cohere-Aya-32B to generate 5 responses, corresponding to scores from 1 to 5 by providing it the rubrics to be used for evaluation using the prompt in Appendix ~\ref{sec:score-gen-prompt}. We use these responses for evaluating our Judge and the score corresponding the response as the ground truth score. We use Cohere Command-A  model and test 6 configurations of LLM-as-a-Judge: (i) Original query along with original response (Baseline) (ii) Original query along with response translated to English (iii) Query translated to English along with original response (iv) Original query and original response along with 2 reference responses as examples (v) Original query and original response along with 4 reference responses as examples (vi) Original query and original response along with 8 reference responses as examples. We note that we only provide  randomly chosen reference responses to the model without any evaluation, making our methodology different from few-shot prompting and eliminating the need for human evaluated responses for reference. As shown in Appendix ~\ref{sec:judge-alignment-ablation}, providing reference examples to the model leads to higher alignment with ground truth scores as evaluated using Pearson correlation and Cohen's Kappa score. Using original query and response along with 8 randomly chosen examples lead to the highest alignment, hence we choose this configuration for our evaluations.

\section{Experiments}
\subsection{Do LLM Responses show quality differences across languages?}
For each question and language pair, we generate 10 responses per model with temperature set to 1. For generating each response, we use the system prompt in Appendix ~\ref{sec:resp-gen-prompt} as the system prompt and the respective query as user prompt. We evaluate all the responses using LLM as a Judge with the temperature set to 0. The system prompt used for evaluating the model is available in Appendix ~\ref{sec:ver-prompt}. Results shown in Figure \ref{fig:model_scores} show that model responses show significant quality differences across languages. Specifically, responses in Hindi, Swahili, and Hebrew are consistently worse than those in English, Chinese, and Brazilian Portuguese. Even Cohere-Aya models, which were trained for multilingual use cases show worse performance in Swahili. We perform the Kruskal–Wallis significance test on the evaluation scores, the results are available in Table ~\ref{tab:kw_model_results}. We find that the p-value is < 0.05 for all models and indicate statistically significant difference in quality of responses.

\begin{table}[h!]
\centering
\begin{tabular}{lccc}
\hline
\textbf{Model} & \textbf{H Statistic} & \textbf{p-value} \\
\hline
aya-32b       & 712.7980  & $8.3941\times10^{-152}$ \\
aya-8b        & 721.1299  & $1.3252\times10^{-153}$\\
magistral-small & 610.8105  & $9.3325\times10^{-130}$\\
qwen3-14b             & 928.9057  & $1.4752\times10^{-198}$\\
sarvam-m                   & 899.8367  & $2.8870\times10^{-192}$ \\
\hline
\end{tabular}
\caption{Kruskal--Wallis test results by model.}
\label{tab:kw_model_results}
\end{table}

\paragraph{}
\label{sec:judge-verification}
To verify that quality differences are not caused by evaluator bias across languages, we translate a subset of English responses to Hindi and a subset of Hindi responses to English. We translate the responses with Gemini-2.5-Flash using temperature set to 0 using the system prompt in the Appendix ~\ref{sec:trans-prompt}. We note that responses originally in English, translated to Hindi, score better than responses originally in Hindi translated to English (Figure ~\ref{sec:judge_translation}). This shows that the responses generated in Hindi are of lower quality and language of the response does not impact the score provided by the LLM Judge. 

\subsection{How does model architecture and training methodology impact answers across languages?}We also note that Cohere-Aya-32B shows smaller performance differences than Cohere-Aya-8b which suggests that larger models show higher consistency across languages. Our results also show that, although both Sarvam-m and Magistral are finetuned variants of Mistral-small-3.1-24B, they perform differently across languages. Sarvam-m provides better responses for English and Hindi while Magistral provides better responses for English, Chinese and Brazilian Portuguese. This suggests that post-training or finetuning can effectively improve model responses for particular languages.

\subsection{Are language and culture entangled?}
\begin{figure}[t]
\centering
\includegraphics[width=\columnwidth]{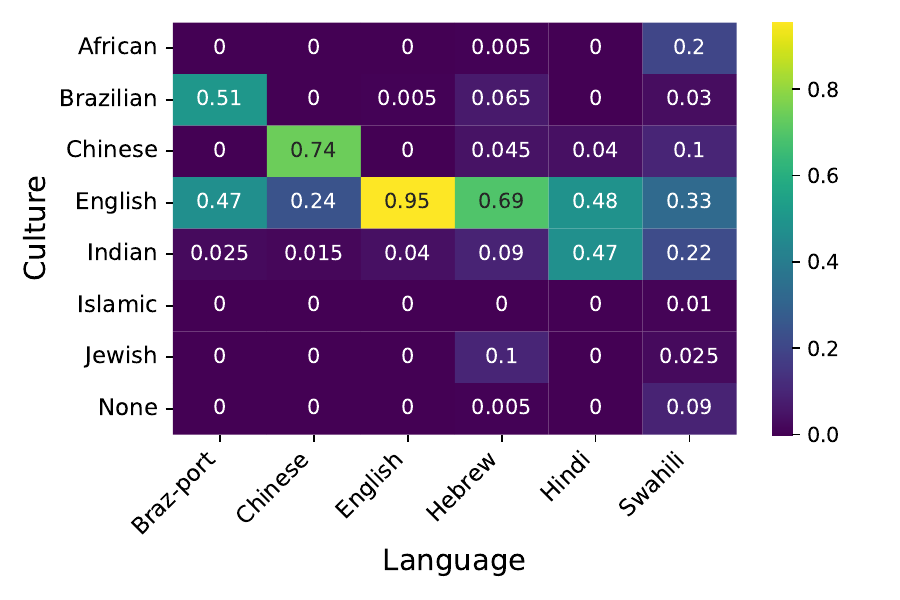}
\caption{\textbf{Results show the proportion of responses classified as each culture by language. X-axis shows the language of the query and Y-axis shows the culture to which the response was classified using LLM as a Judge. }}
\label{fig:culture_heatmap}
\end{figure}

To verify if language and culture are entangled, we translate all the responses from non-English languages to English. We classify each response as one of English/Western, Indian, Chinese, African, Latin American or Jewish culture. We classify each question and response pair using LLM as a Judge with the system prompt in Appendix ~\ref{sec:classify-prompt} and temperature set to 0. The results in Figure \ref{fig:culture_heatmap} show that even after translating all responses to English, the LLM as a Judge is able to classify most answers to the cultural context related to the language in which they were generated. This shows that the answers were not just of lower quality but used different cultural context. This shows that using a language leads to answers with cultural context related to that language.

\begin{figure}[t]
\centering
\includegraphics[width=\columnwidth]{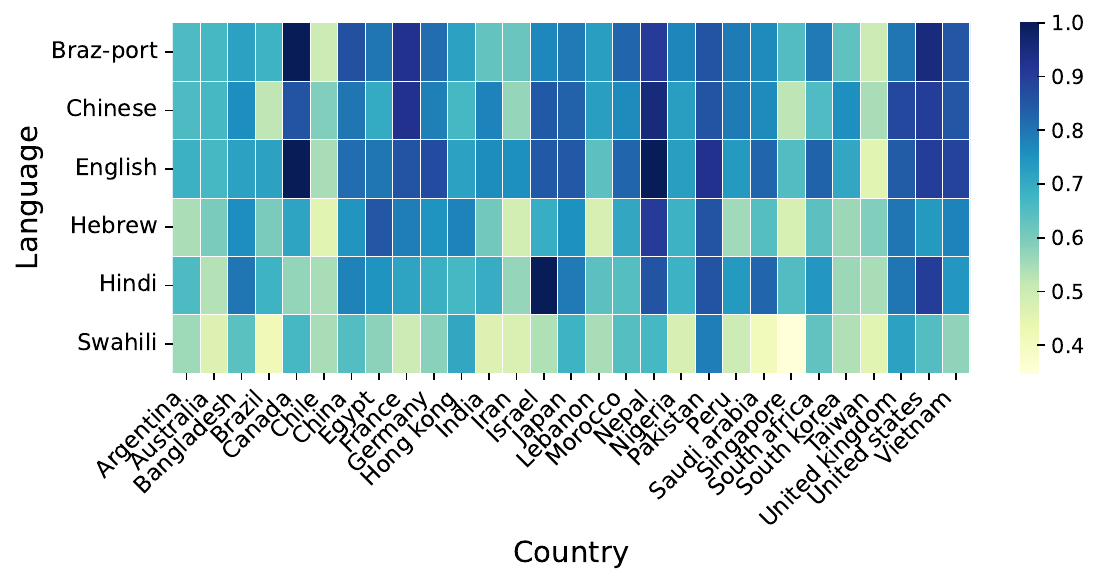}
\caption{\textbf{Accuracy on translated subset of CulturalBench by language and country for Qwen3-14b}}
\label{fig:culturalbench_accuracy}
\end{figure}

To further verify this, we translated a subset of the CulturalBench dataset to Hindi, Chinese, Brazilian Portuguese, Swahili and Hebrew. We evaluated Qwen3-14b on all the translated and the original English version of the benchmark with temperature set to 0. Results show that across languages, the accuracy on cultural questions related to each country varies significantly (Figure \ref{fig:culturalbench_accuracy}). We hypothesize that the relation between language and culture depends on the amount of pretraining data about a particular culture available in that language and the overall representation of that language in the training data. To ensure that the performance difference across languages are significant, we perform the Kruskal-Wallis test. We find that H = 45.5158 and p = $1.1395\times10^{-8}$ verifying that the performance across languages is statistically different. To further verify that the differences are not due to random perturbations, we perform an experiment evaluating the performance on CulturalBench by appending random strings to the queries, based on \cite{mukherjee2024culturalconditioningplaceboeffectiveness} showcasing that adding random strings to queries can lead to similar variations as cultural prompting. We note that while addition of random strings leads to reduced performance, different strings do not lead to variations in performance similar to performance difference across languages. We also perform the Kruskal-Wallis test to evaluate differences across random strings, and find that H = 1.0228 and p = $7.9574\times10^{-1}$. This verifies that performance changes across languages are more significant than random perturbations in the query.

\section{Limitations}
Our work is limited to using LLM as a Judge for evaluating responses. The model being used as LLM Judge may be biased in its evaluation which can affect the results of the study. To mitigate any evaluation bias across languages, we took careful measures and results in subsection ~\ref{sec:judge-verification} shows the robustness of our judge in evaluating responses across languages. Our work is limited to small to moderate open-source models and can be extended to larger models. Further research can also include use of mechanistic interpretability techniques to study the relationship between language and culture in LLMs.

\section{Conclusion}
We demonstrate that for open-ended queries, LLMs provide answers with varying quality and cultural context across languages. We also demonstrate that LLM responses use different cultural context when asked in different languages, which leads to changes in performance on cultural knowledge benchmarks and also impacts the responses for open-ended questions. These results together show a relation between language and culture for LLMs. We call for improved multilingual training data and training methods to increase uniformity in response quality across languages. We urge further research to identify similar biases that negatively affect groups based on language and to develop methods for mitigating such biases.

\bibliographystyle{plainnat}
\bibliography{bibliography}

\clearpage
\appendix
\onecolumn
\section{Questions}
\label{sec:questions}
\begin{table}[H]
\centering
\footnotesize
\setlength{\tabcolsep}{8pt}
\renewcommand{\arraystretch}{1.18}

\begin{tabular}{|p{0.22\textwidth}|p{0.72\textwidth}|}
\hline
\begin{minipage}[t]{\linewidth}\raggedright\textbf{Category}\end{minipage} &
\begin{minipage}[t]{\linewidth}\raggedright\textbf{Queries}\end{minipage} \\
\hline

\begin{minipage}[t]{\linewidth}\raggedright Programming Advice\end{minipage} &
\begin{minipage}[t]{\linewidth}\raggedright
\begin{itemize}
\item I want to learn \{programming language\}, can you suggest a plan to start with?
\item How do I master software engineering and system design concepts?
\end{itemize}
\end{minipage}
\\
\hline

\begin{minipage}[t]{\linewidth}\raggedright Research Advice\end{minipage} &
\begin{minipage}[t]{\linewidth}\raggedright
\begin{itemize}
\item Give me tips and guidelines for writing a good research paper.
\item I am a beginner in \{field\} research. Give me ideas for research problems to work on. Provide ideas along with existing research for reading further.
\end{itemize}
\end{minipage}
\\
\hline

\begin{minipage}[t]{\linewidth}\raggedright Trading/Investing\end{minipage} &
\begin{minipage}[t]{\linewidth}\raggedright
\begin{itemize}
\item What is the best way to do day trading from 100 dollars?
\item I want to buy a house which is costing me 75lacks and my monthly earning is 50k. Do you have any suggestions for this?
\item I want to invest my retirement savings. I have \{amount\}. How do I split my investment across equity, real estate, gold, debt?
\end{itemize}
\end{minipage}
\\
\hline

\begin{minipage}[t]{\linewidth}\raggedright Learning\end{minipage} &
\begin{minipage}[t]{\linewidth}\raggedright
\begin{itemize}
\item Give me a study plan to learn \{language\}.
\end{itemize}
\end{minipage}
\\
\hline

\begin{minipage}[t]{\linewidth}\raggedright Business/Marketing\end{minipage} &
\begin{minipage}[t]{\linewidth}\raggedright
\begin{itemize}
\item I'm looking for a comprehensive business plan to launch a new venture selling printed shirt designs. This plan should detail the initial setup steps and provide a creative, step-by-step social media and Instagram strategy to help me become a star in this industry.
\item Give me 5 tricks about digital marketing on instagram.
\item Act as a business analyst, brainstorm for me 5 novel business ideas that are able to form a startup company using NLP technology.
\item What are the 10 fastest niches in tech growing fast you would focus on to look for problems to solve and start a startup in?
\item What are some ways to make money as a 13 year old from home?
\end{itemize}
\end{minipage}
\\
\hline

\begin{minipage}[t]{\linewidth}\raggedright Job/Interview\end{minipage} &
\begin{minipage}[t]{\linewidth}\raggedright
\begin{itemize}
\item Provide me interview questions and answers for \{job role\} to help me prepare for the interview.
\end{itemize}
\end{minipage}
\\
\hline

\begin{minipage}[t]{\linewidth}\raggedright Health/Medicine\end{minipage} &
\begin{minipage}[t]{\linewidth}\raggedright
\begin{itemize}
\item Suggest home remedies for \{issue\}.
\item How do I improve the quality and duration of my sleep? Give tips and a schedule to follow.
\item Over the past few years, I've noticed a significant decline in my memory. This concerns me, and I worry about the possibility of developing conditions like madness or Alzheimer's in the future. I'm seeking advice on how to improve my overall well-being, specifically my physical, brain, and psychological health.
\item Make me a meal plan of a week's worth of meals. I need 3 meals a day. I must hit a protein goal of 120 grams of protein everyday, and my calories for each day is 3000 calories. Be as detailed as possible and include the calorie count and protein count for each meal. I don't eat meat. Create a grocery list as well.
\item Give me a workout routine for calisthenics with the split push pull legs upper lower workout for beginner and with just dumbbells 5--10 kg, pull-up bar, resistance band, bodyweight, gymnastic rings, and best calories intake for 76kg and 173cm tall.
\item Create a 30 minute work out routine for beginners that I can do at home.
\end{itemize}
\end{minipage}
\\
\hline

\end{tabular}

\caption{Queries by category}
\label{tab:categories_queries}
\end{table}

\section{Score-Wise Responses Generation Prompt}
\label{sec:score-gen-prompt}
\begin{verbatim}
You are a helpful data generation assistant. You will be given a user query.
Your task is to generate multiple responses to the query, 
each evaluated across specific rubrics. 
Each rubric score ranges from 1 (lowest quality) to 5 (highest quality).  

### Rubrics for Evaluation
1. **Detail and Completeness**  
   - Score 1: The response is extremely brief, incomplete, abrupt, 
   or misses most of the requirements.  
   - Score 5: The response is thorough, well-structured, 
   and fully addresses all aspects of the query with rich detail.  

2. **Linguistic Quality**  
   - Score 1: The response has poor grammar, unclear phrasing, 
   awkward sentence construction, or confusing vocabulary.  
   - Score 5: The response is fluent, grammatically correct, clear, 
   and uses precise vocabulary appropriate for the context.  

3. **Factual Correctness**  
   - Score 1: The response contains clear inaccuracies, fabricated information, 
   or misleading claims.  
   - Score 5: The response is entirely accurate, factually reliable, and free of errors.  

4. **Actionability (if advice or steps are requested)**  
   - Score 1: The response is vague, impractical, or does not provide usable steps.  
   - Score 5: The response is highly actionable, offering realistic, clear, 
   and practical guidance that can be implemented easily.  

5. **Relevance to the Query**  
   - Score 1: The response does not address the actual question, goes off-topic, 
   or provides generic advice unrelated to the query.  
   - Score 5: The response directly answers the query, stays on-topic throughout, 
   and avoids unnecessary digressions.  

### Task Instructions
- You must generate **five distinct responses** to the same query.  
- Each response should correspond to a different overall quality level, 
from **1 (lowest)** to **5 (highest)**.  
- All rubric scores for a single response must align with that overall score.  
  - For example, a response at overall level 2 should reflect level 2 in all five rubrics.  
- The quality of responses should progress gradually from poor (score 1) 
to excellent (score 5).  

### Output Format
Your final output must follow this structure in {language}:  
[Response for score 1, Response for score 2, Response for score 3, 
Response for score 4, Response for score 5]
\end{verbatim}

\section{Judge Alignment Ablation}
\label{sec:judge-alignment-ablation}
\begin{figure}[H]
\centering
\includegraphics[width=\textwidth]{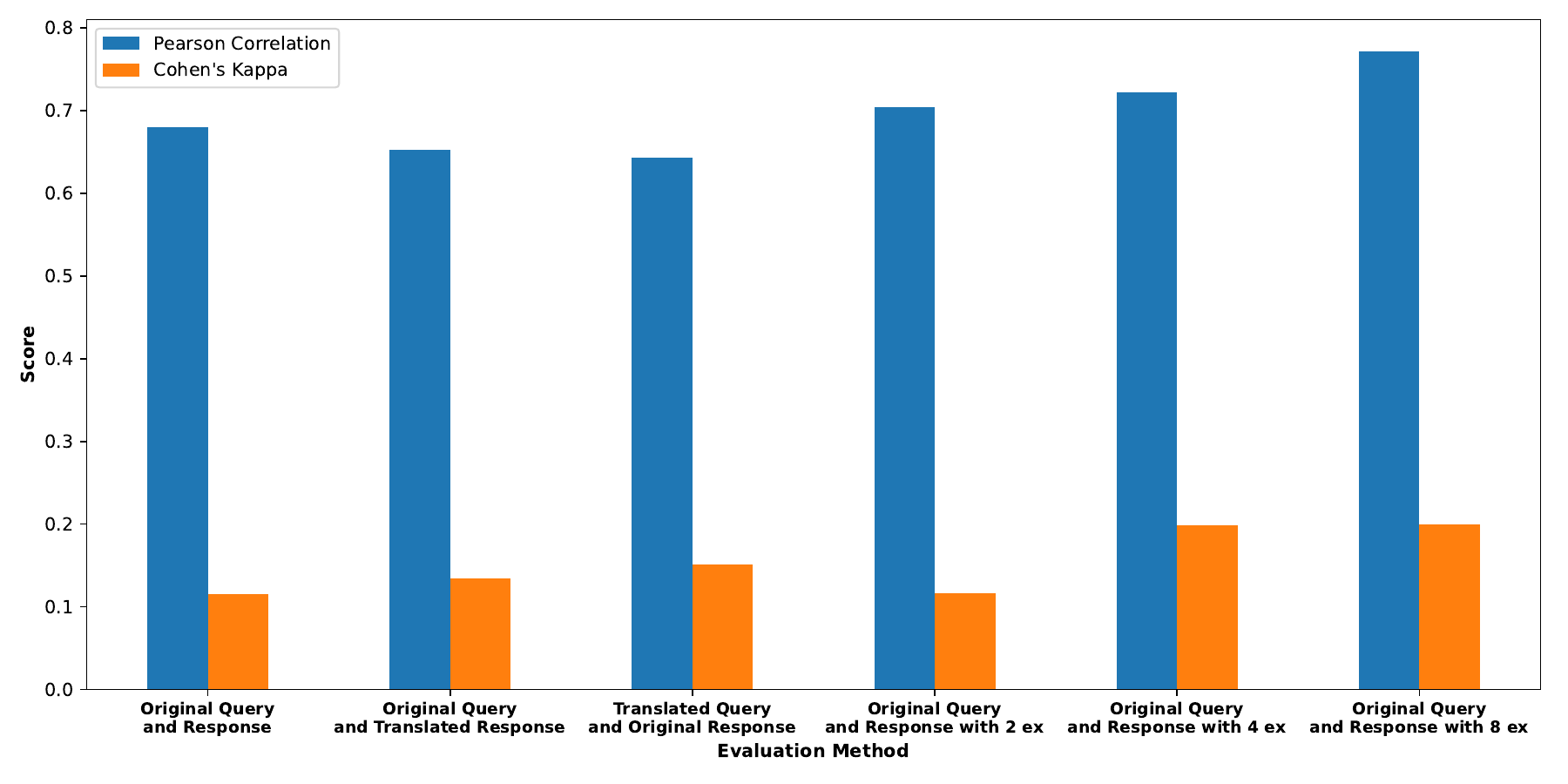}
\caption{\textbf{Comparison of different LLM-as-a-Judge configurations, evaluated on the basis of alignment with ground truth scores in terms of pearson correlation and Cohen's Kappa score. We note that addition of examples (denoted by 'ex' in the graph) lead to higher alignment, with 8 examples resulting in the highest alignment.}}
\label{fig:judge_configurations}
\end{figure}

\section{Response Generation Prompt}
\label{sec:resp-gen-prompt}
\begin{center}
\begin{verbatim}
You are a helpful assistant. You will be
provided a query by the user.
Your task is to go through the query,
understand it and provide appropriate
answer to the user.
Provide your answer in {language}
\end{verbatim}
\end{center}

\section{Verification Prompt}
\label{sec:ver-prompt}
\begin{center}
\begin{verbatim}
You are a helpful evaluation assistant. You
will be given a user query and a
model-generated responseand few example
responses.
Your task is to evaluate the quality of the
response based on specific rubrics. Each
rubric is scored from 1 (lowest) to 5 (highest).

### Rubrics for Evaluation
1. Completeness - Score 1: The response is extremely brief,
incomplete, abrupt, or misses most of
the requirements. 
Score 5: The response is thorough, well-structured, and fully addresses 
all aspects of the query.
2. Linguistic Quality - Score 1: The response has poor grammar,
unclear phrasing, awkward sentence
construction, or confusing vocabulary.
- Score 5: The response is fluent,
grammatically correct, clear, and uses
precise vocabulary appropriate for the
context.
3. Factual Correctness - Score 1: The response contains clear
inaccuracies, fabricated information,
or misleading claims.
- Score 5: The response is entirely accurate,
factually reliable, and free of errors.
4. Actionability (if advice or steps are requested) - Score 1: The response is vague,
impractical, or does not provide usable steps.
- Score 5: The response is highly actionable,
offering realistic, clear, and practical
guidance that can be implemented easily.
5. Riskiness - Score 1: The advice is highly risky,
unsafe, or involves impractical or
dangerous steps.
- Score 5: The advice is low-risk, safe,
and reasonable to implement with minimal
downside.

### Task Instructions
- Assign a score from 1-5 for each rubric.
- Provide an overall score (1-5) that reflects the general quality of the
response across all rubrics.
- Justify each score with a brief explanation
(2-4 sentences), highlighting strengths
and weaknesses.

First go through the query to understand the
requirements of the user/query.
Then go throught the example responses to gain
an idea of what possible responses to the query
could have been like.
Based on the query and the example response,
analyse the model response and score it accordingly.
Keep in mind the example responses are not ideal
or reference responses but examples to get an
idea of what possible responses for the query
could have been.

### Output Format
Your final output must be structured as follows
between the tags correctly:
<evaluation>
{
"Detail and Completeness": {"score": X, "justification": "..."},
"Linguistic Quality": {"score": X, "justification": "..."},
"Factual Correctness": {"score": X, "justification": "..."},
"Actionability": {"score": X, "justification": "..."},
"Riskiness": {"score": X, "justification": "..."},
"Overall": {"score": X, "justification": "..."
}
}
</evaluation>
\end{verbatim}
\end{center}

\section{Answer Translation Prompt}
\label{sec:trans-prompt}
\begin{verbatim}
You are a helpful translation assistant.
You will be provided a text. Your task is to
translate it into {language} while maintaining
the text structure and meaning.
Only translate the text, do not change its
content, meaning or structure in any manner.
Put the response between <sentence></sentence>.
\end{verbatim}

\section{Culture Classification Prompt}
\label{sec:classify-prompt}
\begin{verbatim}
You are an expert in cross-cultural
communication and linguistic anthropology.
Your task is to identify the most likely
cultural context reflected in a given
LLM-generated answer.

You will be given:
1. A user query (the prompt given to the model)
2. The model’s generated response

Each response reflects cultural patterns such as
values, tone, communication style, worldview,
and moral reasoning.
The possible cultures are:
- English (Western/Anglo-American)
- Chinese
- Indian
- Brazilian/Portuguese (Latin)
- African
- Jewish

Analyze the text carefully and decide
which culture the response most likely reflects.
Base your decision on cultural markers such as:
- Emphasis on individualism vs. collectivism
- Formality, respect for hierarchy, or social harmony
- References to religion, family, tradition, or community
- Tone (emotional, rational, moral, pragmatic, etc.)
- Value orientations (e.g., independence, respect, duty, faith, self-expression)
- Common idioms, metaphors, or linguistic patterns

Provide your answer in the following format:
<evaluation>
<culture>one of: English, Chinese,
 Indian, Brazilian/Portuguese,
African, Jewish</culture>
<reason>brief explanation of why this
culture fits best based on linguistic
and cultural cues</reason>
</evaluation>
\end{verbatim}

\section{Judge Translation Ablation}
\label{sec:judge_translation}

\includegraphics[width=\columnwidth]{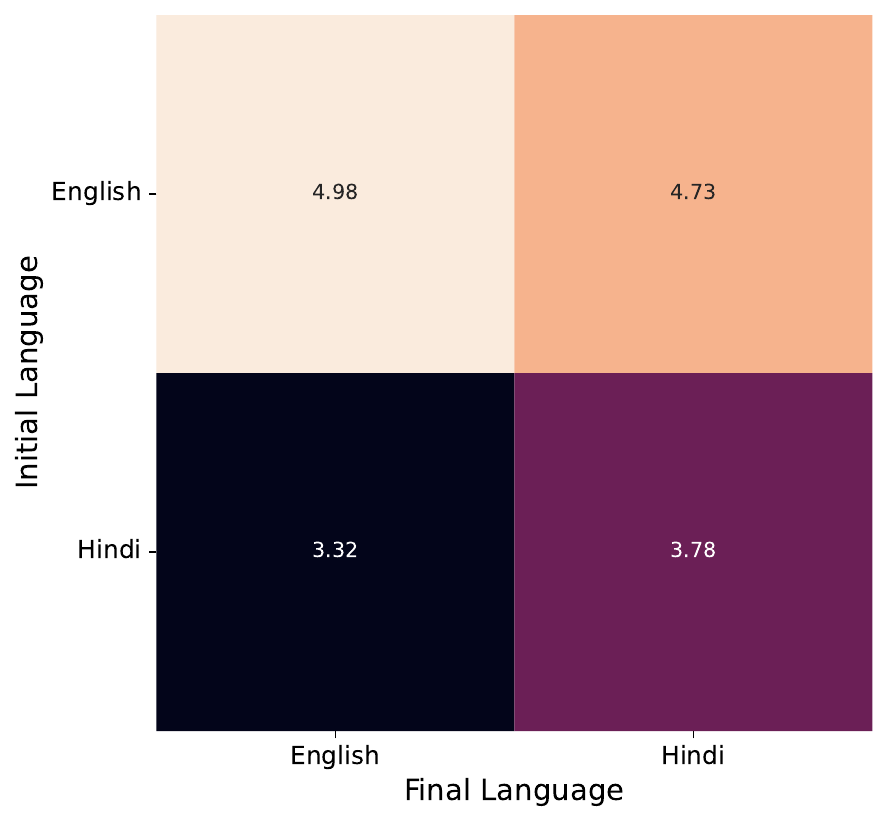}
\captionof{figure}{\textbf{Results comparing raw and translated answers. Y-axis represents the language in which the response was generated. X-axis represents the final language in which the response was provided to LLM as a Judge. Values show the mean score for that Initial and Final language pair.}}
\label{fig:judge_translation}

\section{CulturalBench Random Strings Ablation}
\label{sec:cult-bench-random-string}
\begin{center}
\includegraphics[width=\columnwidth]{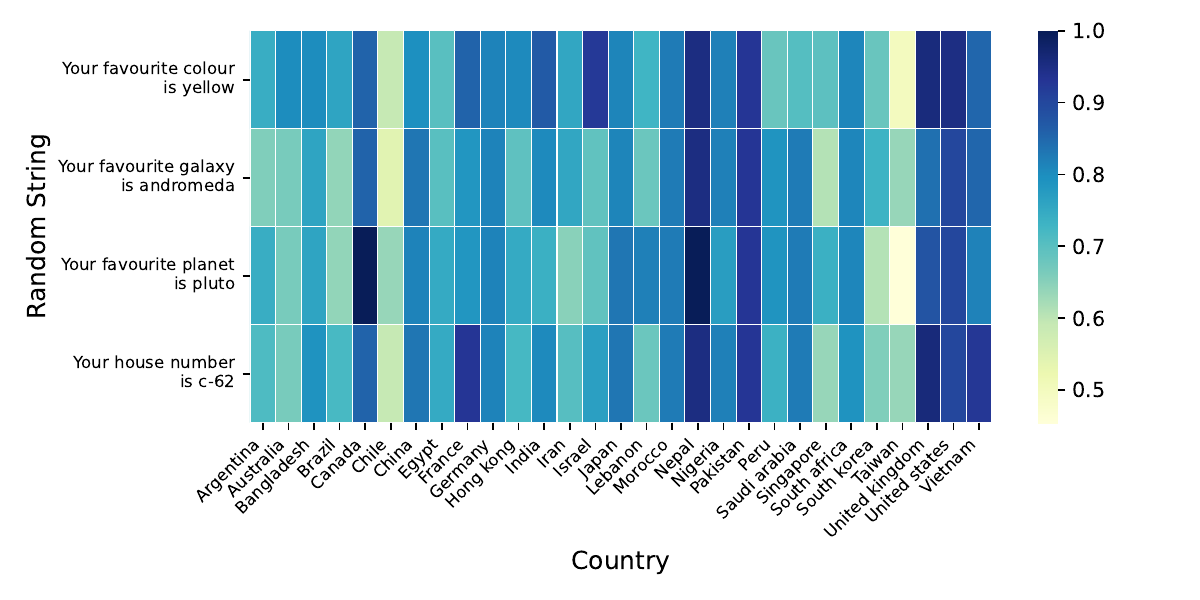}
\captionof{figure}{\textbf{Accuracy of Qwen3-14b on subset of CulturalBench with addition of random strings by country}}
\label{fig:cult-bench-rand-strings}
\end{center}

\clearpage

\end{document}